\documentclass[letterpaper, 10 pt, conference]{ieeeconf}  % Comment this line out if you need a4paper

\IEEEoverridecommandlockouts  % This command is only needed if you want to use the \thanks command

\usepackage{graphics} % for pdf, bitmapped graphics files
\usepackage{epsfig} % for postscript graphics files
\usepackage{mathptmx} % assumes new font selection scheme installed
\usepackage{times} % assumes new font selection scheme installed
\usepackage{amsmath} % assumes amsmath package installed
\usepackage{amssymb}  % assumes amsmath package installed
\usepackage{algpseudocode}
\usepackage{algorithm}
\usepackage{mathtools}
\usepackage[noadjust]{cite}

\usepackage{subcaption}
\usepackage[font=footnotesize]{caption} 
\usepackage{hyperref}
\newcommand{\ours}{$\mathrm{D}^3\mathrm{ROC}$}

\title{\LARGE \bf
Data-Driven Distributionally Robust Optimal Control with State-Dependent Noise
}

\author{Rui Liu, Guangyao Shi, Pratap Tokekar   % <-this % stops a space
\thanks{All authors are from the University of Maryland, College Park, MD 20742 USA. Emails: \tt\small \{ruiliu, gyshi, tokekar\}@umd.edu}%
\thanks{This work is supported in part by the National Science Foundation Grant No. 1943368 and the Army Grant No. W911NF2120076.}
}

\begin{document}
\maketitle
\thispagestyle{empty}
\pagestyle{empty}

%%%%%%%%%%%%%%%%%%%%%%%%%%%%%%%%%%%%%%%%%%%%%%%%%%%%%%%%%%%%%%%%%%%%%%%%%%%%%%%%
\begin{abstract}
Distributionally Robust Optimal Control (DROC) is a framework that enables robust control in a stochastic setting where the true disturbance distribution is unknown. Traditional DROC approaches require given ambiguity sets and KL divergence bounds to represent the distributional uncertainty; however, these quantities are often unavailable a priori or require manual specification. To overcome this limitation, we propose a data-driven approach that jointly estimates the uncertainty distribution and the corresponding KL divergence bound, which we refer to as \ours. To evaluate the effectiveness of our approach, we consider a car-like robot navigation task with unknown noise distributions. The experimental results show that \ours~yields robust and effective control policies, outperforming iterative Linear Quadratic Gaussian (iLQG) control and demonstrating strong adaptability to varying noise distributions.
% Moreover, it shows the effectiveness of our proposed approach in handling different noise distributions. Overall, the proposed approach offers a promising solution to real-world DROC problems where the noise distribution and KL divergence bounds may not be known a priori, increasing the practicality and applicability of the DROC framework.
\end{abstract}
% It integrates the robustness of DRO with the predictive capability of model predictive control to provide robust control policies that can handle uncertain and changing environments

%%%%%%%%%%%%%%%%%%%%%%%%%%%%%%%%%%%%%%%%%%%%%%%%%%%%%%%%%%%%%%%%%%%%%%%%%%%%%%%%
\section{Introduction}
The objective of optimal control \cite{doi:https://doi.org/10.1002/9781118122631.ch2} is to determine the optimal control inputs and state trajectories for a dynamical system that minimize or maximize a pre-defined objective. Most real-world systems are subject to uncertainties \cite{SANJARI2020118525, liu2024towards, liu2024adaptive, liu2025aukt} and robustness concerns \cite{liu2025caml, liu2025imrl, liu2025mmcd}, stemming from factors such as measurement noise, inaccurate dynamics models, and environmental disturbances. To address these challenges, two prominent approaches have been developed: stochastic control \cite{MESBAH2018107, doi:10.1137/140979940} and robust control \cite{lee2017gp, wang2015approximate}. Stochastic control incorporates probabilistic models of uncertainty into the control design process. The underlying distribution of the uncertainty must be known in order for the control design to be effective. Robust control provides robustness to a wide range of uncertainties by considering worst-case scenarios. Both approaches provide practical solutions for controlling dynamical systems in the presence of uncertainty. However, both approaches have limitations. For stochastic control, the probability distribution of uncertainty may not always be available or may be difficult to determine. For robust control, considering worst-case scenarios can sometimes lead to overly conservative control designs that are not necessary and even sub-optimal.

The limitations of stochastic and robust control methods have led to the development of Distributionally Robust Optimal Control (DROC) \cite{nishimura2021rat,coulson2021distributionally}, a rapidly advancing area in control theory that addresses uncertainty in both the system model and the process noise. It merges the robustness properties of Distributionally Robust Optimization (DRO) \cite{rahimian2019distributionally, delage2010distributionally} with the predictive capabilities of Model Predictive Control (MPC) \cite{camacho2013model, kouvaritakis2016model}. It can provide control policies that are robust to uncertain and changing conditions. Unlike traditional stochastic control methods, DROC does not assume prior knowledge of the underlying distributions of the uncertainty. Instead, it assumes there is an ambiguity set of probability measures $\mathbb{P}$ that contain the true distribution, and this $\mathbb{P}$ should be as small as possible \cite{rahimian2019distributionally}. 

% In the field of DRO, there are typically two categories of methods for constructing the ambiguity set: moment-based and discrepancy-based \cite{rahimian2019distributionally}. Moment-based ambiguity sets consist of distributions whose moments have certain characteristics, while discrepancy-based ambiguity sets contain distributions that are similar to a reference distribution according to a specific discrepancy measure. In this work, we focus on discrepancy-based ambiguity sets, more specifically the Kullback-Leibler (KL) divergence measure \cite{hu2013kullback}. Since it is a popular measure and provides a way to quantify the uncertainty in the noise distribution.

% This is because we first need to learn the reference distribution as a Gaussian in our approach. KL divergence \cite{joyce2011kullback} is a popular measure of the difference between two probability distributions and is used in DRO to quantify the similarity between the reference distribution and the true distributions in the ambiguity set. It provides a way to measure the uncertainty in the noise distribution, which can be used to determine the distributionally robust control policy that accounts for this uncertainty. In this way, DROC offers a powerful solution for controlling dynamic systems in the presence of uncertainty and changing conditions, making it a valuable approach.

A common way of modeling the ambiguity set is by assuming a reference distribution such that the true distribution is within a given KL divergence bound of this reference distribution \cite{nishimura2021rat, hu2013kullback, li2018kullback, duchi2018learning}. However, in practical applications, it may not be possible to obtain these values in advance. Therefore, in this paper, we aim to address the following question: \emph{Can we learn the reference distribution and KL divergence bound to enhance the effectiveness of DROC?} 
% This poses a research challenge since the accuracy and robustness of the control performance depend on the quality of the estimates for the reference distribution and KL divergence bound. 
% , ensuring the robustness and performance of the control policies.
% To address this challenge, two different scenarios are considered: (1) the true probability distribution of the noise is stationary, and (2) the true probability distribution of the noise is variable, such as changing with the state. Both scenarios present difficulties for estimating the reference distribution and KL divergence bound, necessitating innovative techniques to obtain reliable and accurate estimates. 

Building on DROC \cite{nishimura2021rat}, we introduce Data-Driven Distributionally Robust Optimal Control ($\mathrm{D}^3\mathrm{ROC}$), which jointly learns the reference distribution and the KL divergence bound of the underlying uncertainty. To this end, we consider both stationary and state-dependent noise distributions, employing Gaussian Process (GP) regression \cite{schulz2018tutorial} to estimate the reference noise distribution. In parallel, we use k-Nearest Neighbor (kNN) \cite{wang2009divergence} to estimate the KL divergence bound. This non-parametric design eliminates the need for prior knowledge of the underlying distributions, enabling reliable estimation of the KL-divergence bound directly from observed data.

We evaluate our approach on a car-like robot navigation task with unknown noise distributions, where the objective is to reach the origin. Experimental results show that \ours~achieves smaller mean final distances to the origin and lower corresponding standard deviations, indicating greater accuracy compared to iLQG. Moreover, under \ours, the robot naturally avoids regions with higher noise variance due to its risk-averse behavior, whereas the risk-neutral iLQG tends to pass through these high-variance areas. These results demonstrate that \ours~produces more robust and effective control policies than iLQG.

\section{Related Work}
\subsection{Model Predictive Control}
In MPC \cite{camacho2013model,kouvaritakis2016model,grune2017nonlinear,mayne2014model}, a finite horizon optimization problem is solved recursively at each time step, and the optimal control inputs for the current time step are applied to the system. MPC can be designed for both linear and nonlinear systems and can handle systems with multiple inputs and outputs, time-varying constraints, and stochastic uncertainties. The core idea behind MPC is to use the mathematical model of the system to predict its future behavior and then optimize the control inputs to achieve a desired objective, such as minimizing energy consumption, maximizing production efficiency, or tracking a reference trajectory. MPC has been applied in a variety of fields, including process control \cite{bequette2003process}, automotive engineering \cite{hrovat2012development}, and robotics \cite{zhang2016learning}.

\subsection{Risk Sensitive Control}
Risk-sensitive control \cite{fleming1995risk, liu2024towards} is a subfield of control theory that deals with risk in dynamic systems. This type of control is motivated by the fact that in many real-world applications, decision makers must balance the trade-off between the reward of a successful control outcome and the cost of failure. So the aim of risk-sensitive control is to design control policies that consider both performance and risk, achieving a desired level of control. Mathematically, risk-sensitive control is equivalent to the inner loop of DRO \cite{petersen2000minimax,nishimura2021rat}. It incorporates a measure of risk, known as an entropic risk measure \cite{nass2019entropic}, into the control design process to ensure robustness in the face of uncertainty and variability in the system. The entropic risk measure is superior to the conventional cost function in that it includes higher-order moments of the cost function by computing the expectation of the exponential of the cost function. For linear systems with quadratic cost functions and Gaussian process noise, closed-form optimal control policies can be obtained \cite{farshidian2015risk,roulet2020convergence,nishimura2021rat}.

\subsection{Distributionally Robust Optimization}
DRO \cite{rahimian2019distributionally,bertsimas2019adaptive,delage2010distributionally,goh2010distributionally,levy2020large} is a mathematical framework for decision-making under uncertainty. Unlike traditional stochastic optimization techniques that assume the uncertainty is precisely known and follows a specific probability distribution, DRO takes into account the fact that the true distribution of the uncertainty is often unknown. Instead, DRO models the uncertainty using a ``worst-case" scenario, where the true distribution is restricted to belong to a given ambiguity set. This approach provides robustness against worst-case scenarios and helps mitigate the risk of making poor decisions due to inaccuracies in the modeling of the uncertainty.
% DRO has found applications in a wide range of fields, including finance \cite{gulpinar2007worst}, engineering \cite{miao2021data}, and operations research \cite{shang2018distributionally}, and has become an important tool for decision making under uncertainty in these domains.

\section{Problem Formulation}
Consider the following general nonlinear stochastic system: 
\begin{equation} \label{dyn}
    \mathbf{x}_{t+1} = \mathbf{f}(\mathbf{x}_t,\mathbf{u}_t) + \mathbf{g}(\mathbf{x}_t,\mathbf{u}_t)\mathbf{w}(\mathbf{x}_t,\mathbf{u}_t), 
\end{equation}
where $\mathbf{x}_t \in \mathrm{X}\subset\mathbb{R}^n$ and $\mathbf{u}_t \in \mathrm{U}\subset\mathbb{R}^m$ denote the state and control of the system at step $t$, respectively. $\mathbf{f}$ is the dynamics model of the system, $\mathbf{g}$ is a mapping function, $\mathbf{w} \in \mathbb{R}^c$ is the process noise with unknown distribution. $\mathbf{w}$ may be stationary or vary with states or control inputs.

Even though the true distribution $p$ of the process noise $\mathbf{w}$ is unknown, we assume it is contained in an ambiguity set $\mathbb{P}$ with reference distribution $q$ \cite{hu2013kullback}. The ambiguity set $\mathbb{P}$ is constructed by:
\begin{equation}
    \mathbb{P} = \{p: {D}(p||q)\leq d\} \ ,
\end{equation}
where $D(\cdot||\cdot)$ is the KL divergence and $d > 0$ is the bound. 
% As mentioned before, KL divergence is a measure of the difference between two probability distributions. It quantifies the similarity between the true distribution and the reference distribution. 

$J$ is the cost function over a finite horizon $n$:
\begin{equation}
    J = \sum_{t=0}^{n-1}l_t(\mathbf{x}_t,\mathbf{u}_t)+l_{f}(\mathbf{x}_{n})\ ,
\end{equation}
where $l_t$ is the stage cost, $l_{f}$ is the terminal cost, we consider linear quadratic regulator:
\begin{equation}
    l_t(\mathbf{x},\mathbf{u}) = \frac{1}{2}\mathbf{x}^\intercal \mathbf{Q}_t\mathbf{x} + \frac{1}{2}\mathbf{u}^\intercal \mathbf{R}_t\mathbf{u} \ ,
\end{equation}
\begin{equation}
    l_{f}(\mathbf{x}) = \frac{1}{2}\mathbf{x}^\intercal \mathbf{Q}_{f}\mathbf{x} \ .
\end{equation}

In formulating the DROC problem, we are interested in finding a control policy that minimizes the worst-case expected value of the cost function $J$: 
\begin{equation}
    \min_{\mathbf{u}\in\mathrm{U}} \max_{p\in\mathbb{P}} \mathbb{E}_p[J] \ .
\end{equation}
According to \cite{hu2013kullback,nishimura2021rat}, the above min-max problem over true distribution $p$ can be converted to the following min-min problem by taking the expectation over the reference distribution $q$:
\begin{equation} \label{droc}
    \min_{\mathbf{\theta}\in\mathrm{\Xi}} \min_{\mathbf{u}\in\mathrm{U}} \biggl\{\frac{1}{\theta}\log[\mathbb{E}_q(e^{\theta J})] + \frac{d}{\theta}\biggr\} \ ,
\end{equation}
where $\theta$ is called the risk-sensitivity parameter, and $\Xi$ is a non-empty set of positive $\theta$ that gives finite entropic risk measure.

To solve this DROC problem, knowledge of the reference distribution $q$ of the noise, and the KL divergence bound $d$, are required. In the next section, we show how to compute $q$ and $d$ in a data-driven fashion when they are unknown. 

% \subsubsection{Stationary Noise Distribution}
% If the true distribution of the noise, denoted by $p$, is stationary and does not change with states, we can estimate the reference distribution $q$ as a stationary distribution. The KL divergence between $p$ and $q$ will also be stationary consequently. This means that the bound, which is used to construct the ambiguity set for the DROC problem, will also be fixed.

% \subsubsection{State-Dependent Noise Distribution} \label{sdnoise}
% If the true distribution of the noise $p$ varies with the system state, the estimated reference distribution and the KL divergence bound will be non-stationary, changing with the system state. To handle this situation, we propose a data-driven approach to learn the non-stationary reference distribution $q$, and aim to estimate a global maximum bound of the KL divergence from the observed data that can bound all the varying true distributions and reference distributions as the system states change.

\section{The \ours~Solution}
In this section, we present \ours. The objective of \ours~is to first estimate the uncertainty distribution and a bound for the KL divergence, and then solve the optimization problem of Eq. \ref{droc}. \ours~distinguishes itself from traditional DROC approaches that rely on given ambiguity sets \cite{nishimura2021rat, hu2013kullback, li2018kullback, duchi2018learning} by utilizing data-driven techniques. In Section \ref{ddp}, we address the inner minimization over control inputs $\mathbf{u}$ using Differential Dynamic Programming (DDP) \cite{tassa2014control, bechtle2020curious, nishimura2021rat}. However, for DDP to be effective, we require the reference distribution $q$, which we estimate using observed data. Next, for the outer minimization over the risk-sensitivity parameter $\theta$, we adopt the cross-entropy method as described in \cite{nishimura2021rat}. To apply this method successfully, we estimate the KL divergence bound $d$. We study two scenarios for the parameters estimation: (1) stationary noise distribution in Section \ref{stationary}, and (2) state-dependent noise distribution in Section \ref{state-dependent}.

\subsection{Differential Dynamic Programming} \label{ddp}
We utilize DDP to handle the inner minimization of Eq. \ref{droc}, which is equivalent to minimizing $\frac{1}{\theta}\log[\mathbb{E}_q(e^{\theta J})]$, defined as $R_{\theta}(J)$, also known as the entropic risk measure \cite{nass2019entropic, nishimura2021rat}. We assume $\mathbf{g}$ in Eq. \ref{dyn} as an identity mapping for simplicity. The process noise $\mathbf{w}$ is estimated as a Gaussian with zero mean and covariance matrix $\mathbf{W}$. We linearly approximate the system dynamics and quadratically approximate the cost function in terms of state and control deviations $\delta \mathbf{x}_t = \mathbf{x}_t - \mathbf{x}_t^{nom}$ and $\delta \mathbf{u}_t = \mathbf{u}_t - \mathbf{u}_t^{nom}$, where $\mathbf{x}_t^{nom}$ and $\mathbf{u}_t^{nom}$ are the nominal trajectories.

The linearized model: 
\begin{equation}
    \delta \mathbf{x}_{t+1} = \mathbf{A}_t\delta \mathbf{x}_t + \mathbf{B}_t\delta \mathbf{u}_t + \mathbf{w}_t \ ,
\end{equation}
the stage cost approximation:
\begin{equation}
\begin{split}
    \Tilde{l}_t(\delta\mathbf{x}_t, \delta\mathbf{u}_t) = q_t &+ \mathbf{q_t}^\intercal \delta \mathbf{x}_t + \mathbf{r_t}^\intercal \delta \mathbf{u}_t + \frac{1}{2}\delta \mathbf{x}_t^\intercal \mathbf{Q}_t\delta \mathbf{x}_t \\
    & + \frac{1}{2}\delta \mathbf{u}_t^\intercal \mathbf{R}_t\delta \mathbf{u}_t\ ,
\end{split}
\end{equation}
and the terminal cost approximation:
\begin{equation}
    \Tilde{l}_f(\delta\mathbf{x}_{n}) = q_{n} + \mathbf{q}_{n}^\intercal \delta \mathbf{x}_{n} +  \frac{1}{2}\delta \mathbf{x}_{n}^\intercal \mathbf{Q}_{n}\delta \mathbf{x}_{n}\ .
\end{equation}

Then apply the principle of optimality, the Bellman equation for solving the optimal value function is: 
\begin{equation} \label{bellman}
    V_t(\delta \mathbf{x}_t) = \min_{\delta \mathbf{u}_t}\left\{\Tilde{l}_t(\delta\mathbf{x}_t,\delta\mathbf{u}_t)+R_{\theta}\left(V_{t+1}(\delta\mathbf{x}_{t+1})\right)\right\}\ ,
\end{equation}
where $R_{\theta}$ is the entropic risk measure. 

Suppose the value function is of quadratic form expressed as: 
\begin{equation} \label{valuef}
    V_t(\delta\mathbf{x}) = \frac{1}{2}\delta \mathbf{x}^\intercal\mathbf{S}_t\delta\mathbf{x} + \mathbf{s}_t^\intercal\delta\mathbf{x} + s_t\ .
\end{equation}
Then, 
\begin{equation} \label{vt+1}
\begin{split}
     & V_{t+1}(\delta\mathbf{x}_{t+1}) \\
     &= \frac{1}{2}\delta\mathbf{x}_{t+1}^\intercal\mathbf{S}_{t+1}\delta\mathbf{x}_{t+1} + \mathbf{s}_{t+1}^\intercal\delta\mathbf{x}_{t+1} + s_{t+1} \\
     &=\frac{1}{2}(\mathbf{A}_t\delta \mathbf{x}_t + \mathbf{B}_t\delta \mathbf{u}_t + \mathbf{w}_t)^\intercal\mathbf{S}_{t+1}(\mathbf{A}_t\delta \mathbf{x}_t + \mathbf{B}_t\delta \mathbf{u}_t + \mathbf{w}_t) + \\
     & \hspace{5mm}\mathbf{s}_{t+1}^\intercal(\mathbf{A}_t\delta \mathbf{x}_t + \mathbf{B}_t\delta \mathbf{u}_t + \mathbf{w}_t) + s_{t+1}\ .
\end{split}
\end{equation}
% where $w$ is the uncertainty, which is modeled as Gaussian distribution with zero mean and covariance matrix $W$, $w\sim\mathcal{N}(\mathbf{0}, W)$.

Let $\mathbf{z}_t = \mathbf{A}_t\delta \mathbf{x}_t + \mathbf{B}_t\delta \mathbf{u}_t + \mathbf{w}_t$, where $\mathbf{A}_t$ and $\mathbf{B}_t$ are Jacobian matrices of the system model. The true distribution of the noise $\mathbf{w}$ is unknown. But we model it as a Gaussian distribution with zero mean and covariance $\mathbf{W}$, then $\mathbf{z}_t\sim\mathcal{N}(\mathbf{A}_t\delta \mathbf{x}_t + \mathbf{B}_t\delta \mathbf{u}_t, \mathbf{W}_t)$. 
Put Eq. \ref{vt+1} into Eq. \ref{bellman}, we have: 
\begin{equation} \label{min_eq}
\begin{split}
    & V_t(\delta\mathbf{x}_t) \\
    &= \min_{\delta\mathbf{u}_t}\biggl\{q_t + \mathbf{q_t}^\intercal \delta \mathbf{x}_t + \mathbf{r_t}^\intercal \delta \mathbf{u}_t + \frac{1}{2}\delta \mathbf{x}_t^\intercal \mathbf{Q}_t\delta \mathbf{x}_t + \frac{1}{2}\delta \mathbf{u}_t^\intercal \mathbf{R}_t\delta \mathbf{u}_t \\ 
    & \hspace{10mm} + \frac{1}{\theta}log\biggl\{\mathbb{E}_{\mathbf{w}_t}\biggl[exp\biggl(\theta \biggl(\frac{1}{2}(\mathbf{A}_t\delta \mathbf{x}_t + \mathbf{B}_t\delta \mathbf{u}_t + \mathbf{w}_t)^\intercal\mathbf{S}_{t+1} \\
    & \hspace{10mm} (\mathbf{A}_t\delta \mathbf{x}_t + \mathbf{B}_t\delta \mathbf{u}_t + \mathbf{w}_t) + \mathbf{s}_{t+1}^\intercal(\mathbf{A}_t\delta \mathbf{x}_t + \mathbf{B}_t\delta \mathbf{u}_t + \mathbf{w}_t) \\
    & \hspace{10mm} + s_{t+1}\biggr)\biggr)\biggr]\biggr\}\biggr\} \\
    &= \min_{\delta\mathbf{u}_t}\biggl\{q_t + \mathbf{q_t}^\intercal \delta \mathbf{x}_t + \mathbf{r_t}^\intercal \delta \mathbf{u}_t + \frac{1}{2}\delta \mathbf{x}_t^\intercal \mathbf{Q}_t\delta \mathbf{x}_t + \frac{1}{2}\delta \mathbf{u}_t^\intercal \mathbf{R}_t\delta \mathbf{u}_t + \\ 
    & \hspace{10mm} \frac{1}{2}(\mathbf{A}_t\delta \mathbf{x}_t + \mathbf{B}_t\delta \mathbf{u}_t)^\intercal\mathbf{S}_{t+1}(\mathbf{A}_t\delta \mathbf{x}_t + \mathbf{B}_t\delta \mathbf{u}_t) \\ 
    & \hspace{10mm} + (\mathbf{A}_t\delta \mathbf{x}_t + \mathbf{B}_t\delta \mathbf{u}_t)^\intercal \mathbf{s}_{t+1} \\
    & \hspace{10mm} + \frac{1}{\theta}log\biggl\{\mathbb{E}_{\mathbf{z}_t}\biggl[exp\biggl(\theta \biggl(\frac{1}{2}\mathbf{z}_t^\intercal \mathbf{S}_{t+1}\mathbf{z}_t + \mathbf{s}_{t+1}^\intercal\mathbf{z}_t\biggr)\biggr)\biggr]\biggr\} \\ & \hspace{10mm} +s_{t+1}\biggr\}\ ,
\end{split}
\end{equation}
where $q_t, \mathbf{q}_t, \mathbf{r}_t, \mathbf{Q}_t, \mathbf{R}_t$ are the Taylor expansion coefficients of the cost function around the nominal trajectory. 

The expectation on the right hand side of Eq. \ref{min_eq} can be calculated using characteristic function of Gaussian distribution. Then we can minimize over $\delta\mathbf{u}$ and get the optimal control policy:

% \begin{equation}
% \begin{split}
%     & \mathbb{E}\left[exp\left(\theta \left(\frac{1}{2}\mathbf{z}_t^\intercal \mathbf{S}_{t+1}\mathbf{z}_t + \mathbf{s}_{t+1}^\intercal\mathbf{z}_t\right)\right)\right] \\ 
%     &= \int p(\mathbf{z}_t)exp\left(\theta \left(\frac{1}{2}\mathbf{z}_t^\intercal \mathbf{S}_{t+1}\mathbf{z}_t + \mathbf{s}_{t+1}^\intercal\mathbf{z}_t\right)\right)d\mathbf{z}_t \\ 
% \end{split}
% \end{equation}
% &= \int \frac{1}{\sqrt{(2\pi)^n|\mathbf{W}_tdt|}}e^{-\frac{1}{2}(\mathbf{z}_t-\Bar{\mathbf{z}_t})^\intercal (\mathbf{W}_tdt)^{-1}(\mathbf{z}_t-\Bar{\mathbf{z}_t})}\cdot \\ 
% where $p(\mathbf{z})$ is the probability density function of $\mathbf{z}$, which is estimated as a Guassian. After geting this expectation, we can minimize over $\delta\mathbf{u}$ and will get the optimal control policy.
% Eventually, minimizing over $\delta\mathbf{u}$, we will get the optimal control policy.
% \begin{equation}
% \begin{split}
%     \delta\mathbf{u}_t &= \mathbf{k}_t + \mathbf{K}_t\delta\mathbf{x}_t \\
%     \mathbf{k}_t &= \\
%     \mathbf{K}_t &=
% \end{split}
% \end{equation}

\begin{equation}
\begin{split}
    \delta\mathbf{u}_t &= \mathbf{k}_t + \mathbf{K}_t\delta\mathbf{x}_t\ , \\
    \mathbf{k}_t &= -\mathbf{H}_t^{-1}\mathbf{g}_t\ , \\
    \mathbf{K}_t &= -\mathbf{H}_t^{-1}\mathbf{G}_t\ ,
\end{split}
\end{equation}
where 
\begin{equation}
\begin{split}
    \mathbf{M}_t &= \mathbf{W}_t^{-1}-\theta\mathbf{S}_{t+1}\ , \\
    \mathbf{H}_t &= \mathbf{R}_t + \mathbf{B}_t^\intercal\bigl(\mathbf{I}+\theta\mathbf{S}_{t+1}\mathbf{M}_t^{-1}\bigr)\mathbf{S}_{t+1}\mathbf{B}_t\ , \\
    \mathbf{G}_t &= \mathbf{B}_t^\intercal\bigl(\mathbf{I}+\theta\mathbf{S}_{t+1}\mathbf{M}_t^{-1}\bigr)\mathbf{S}_{t+1}\mathbf{A}_t\ , \\
    \mathbf{g}_t &= \mathbf{r}_t + \mathbf{B}_t^\intercal\bigl(\mathbf{I}+\theta\mathbf{S}_{t+1}\mathbf{M}_t^{-1}\bigr)\mathbf{s}_{t+1}\ . \\
\end{split}
\end{equation}

The backward recursions are: 
\begin{equation}
    \begin{split}
    \mathbf{S}_t &= \mathbf{Q}_t + \mathbf{A}_t^\intercal\bigl(\mathbf{I} + \theta\mathbf{S}_{t+1}\mathbf{M}^{-1}\bigr)\mathbf{S}_{t+1}\mathbf{A}_t \\
    & \hspace{10mm} + \mathbf{K}_t^\intercal\mathbf{H}_t\mathbf{K}_t + \mathbf{K}_t^\intercal\mathbf{G}_t + \mathbf{G}_t^\intercal\mathbf{K}_t\ , \\
    \mathbf{s}_t &= \mathbf{q}_t + \mathbf{A}_t^\intercal\bigl(\mathbf{I} + \theta\mathbf{S}_{t+1}\mathbf{M}_t^{-1}\bigr)\mathbf{s}_{t+1} \\
    & \hspace{10mm} + \mathbf{K}_t^\intercal\mathbf{H}_t\mathbf{k}_t + \mathbf{K}_t^\intercal\mathbf{g}_t + \mathbf{G}_t^\intercal\mathbf{k}_t\ , \\
    s_t &= q_t + s_{t+1} - \frac{1}{2\theta}\log\bigl(\det(\mathbf{I} - \theta\mathbf{W}_t\mathbf{S}_{t+1})\bigr) \\
    & \hspace{10mm} + \frac{\theta}{2}\mathbf{s}_{t+1}^\intercal\mathbf{M}_t^{-1}\mathbf{s}_{t+1} + \frac{1}{2}\mathbf{k}_t^\intercal\mathbf{H}_t\mathbf{k}_t + \mathbf{k}_t^\intercal\mathbf{g}_t\ ,
    \end{split}
\end{equation}
for $t=n$ goes backward to 0, with initial conditions $s_{n}=q_{n}, \mathbf{s}_{n}=\mathbf{q}_{n}, \mathbf{S}_{n} = \mathbf{Q}_{n}$.
% The proof can refer to \cite{bechtle2020curious,farshidian2015risk}. 

After addressing the inner minimization using DDP, the outer minimization is solved using the cross-entropy method, following the approach described in \cite{nishimura2021rat}. However, for both DDP and cross-entropy method to be effective, it is required to estimate the noise reference distribution and the KL divergence bound.

We will focus on the next sections to address these requirements. Firstly, we examine the case of a stationary noise distribution in section \ref{stationary}. Then, we investigate the case of a state-dependent noise distribution in section \ref{state-dependent}.

\subsection{Stationary Noise Distribution} \label{stationary}
\subsubsection{Estimating Reference Distribution} \label{staq}
We employ Maximum likelihood estimation (MLE) for estimating the parameters of a probability distribution based on observed data. Once we have estimated the parameters of $q$, we can use them to construct the ambiguity set for the DROC problem.
\subsubsection{Estimating KL Divergence Bound} \label{stad}
We utilize k Nearest Neighbor (kNN) \cite{wang2009divergence} to estimate the KL divergence bound. This method is based on the assumption that the KL divergence between two distributions can be estimated from their samples. Euclidean distance is used to measure the distance between samples. According to \cite{wang2009divergence}, the estimated KL divergence $\hat{D}(p||q)$ between distributions $p$ and $q$ can be written as:
\begin{equation}
\hat{D}(p \| q)=\frac{r}{N} \sum_{i=1}^N \ln \frac{\nu_i}{\rho_i}+\log \frac{M}{N-1}\ ,
\end{equation}
where $r$ is the dimension of the data, $N$ is the number of samples drawn i.i.d from distribution $p$, $M$ is the number of samples drawn i.i.d from distribution $q$. $\rho_i$ is the distance between $i$-th element drawn from $p$ and its k-th nearest neighbor in samples drawn from $p$ except itself, $\nu_i$ is the distance between $i$-th element drawn from $p$ and its k-th nearest neighbor in samples drawn from $q$.

% Zhao and Lai \cite{9142281} provided a bound for the bias of the estimated KL divergence
% \begin{equation}
% |\mathbb{E}[\hat{D}(p \| q)]-D(p \| q)|= O\left(\left(\frac{\log \min \{M, N\}}{\min \{M, N\}}\right)^{\frac{1}{r}}\right)
% \end{equation}

Once the reference distribution $q$ is estimated as a Gaussian, $M$ samples can be drawn from $q$. The KL divergence bound can then be estimated using kNN estimation with these samples and $N$ true noise samples.

\subsection{State-Dependent Noise Distribution}\label{state-dependent}
\subsubsection{Estimating Reference Distribution} \label{nonstaq}
In cases where the true distribution $p$ of the noise is non-stationary and varies with the system's state, we can learn the reference distribution $q$ of the noise from observed data, as shown in Algorithm \ref{algo1}. For each training state $\mathbf{\Bar{x}}_j$ (${j = 1, \ 2, \cdots, \ m}$), $N$ true noise samples $\{\mathbf{w}_j^{(1)},\ \mathbf{w}_j^{(2)},\ \cdots,\ \mathbf{w}_j^{(N)}\}$ are available. We assume that the noise of each dimension does not impact other dimensions for simplicity. Consequently, we can separate each dimension of the state and its corresponding noise. Then, we employ MLE to estimate the variance of the noise for each dimension. For instance, when considering the $x$ coordinate of the state, we can obtain a set of tuples $\{(\Bar{x}_1,\ v_1),\ (\Bar{x}_2,\ v_2),\ \cdots,\  (\Bar{x}_m,\ v_m)\}$, where $v$ denotes the variance for the $x$ coordinate. 

We train a Gaussian Process (GP) estimator \cite{schulz2018tutorial} for each dimension of the state to learn the reference distribution. We use the set $\{(\mathbf{\Bar{x}}_1^{(i)},\ v_1^{(i)}),\ (\mathbf{\Bar{x}}_2^{(i)},\ v_2^{(i)}),\ \cdots,\ (\mathbf{\Bar{x}}_m^{(i)},\ v_m^{(i)})\}$, where $\mathbf{\Bar{x}}_j^{(i)}$ denotes the $i$th dimension of the $j$th training state and $v_j^{(i)}$ denotes the corresponding variance of the noise. We use zero mean and squared-exponential kernel function for GP. The kernel function $k(\cdot,\ \cdot)$ is defined as $k(a,a') = \sigma^2\exp\left(-\frac{(a-a')^2}{2l^2}\right)$, where $\sigma^2$ is the signal variance, $l$ is the length scale. The hyperparameters signal variance and length scale are learned by maximizing the log-likelihood of the training data.

When given a new state $\mathbf{x}$, we use the trained GPs to predict the variances of the noise for each dimension, denoted as $\{\hat{v}_1,\ \hat{v}_2,\ \dots,\ \hat{v}_r\}$, where $r$ is the number of dimensions. These predicted variances are then combined into a diagonal covariance matrix, which represents the reference distribution $q$ for the given state.

 % In summary, the key idea behind Algorithm \ref{algo1} is to use GPs to model the reference distribution $q$ of the noise, which varies with the system's states. This is achieved by training a separate GP for each dimension of the state, using the variance of the noise obtained from MLE as the target variable. This approach enables us to predict the variance for each dimension of the noise given a new state, and then combine these variances into a diagonal covariance matrix to represent the reference distribution $q$.

 \begin{algorithm}
\caption{Estimation of reference distribution for state-dependent noise}\label{algo1}
\textbf{Input:} State $\mathbf{x}$ of the system \\
\textbf{Output:} State-dependent reference distribution $q(\mathbf{x})$
\begin{algorithmic}[1]
\Require Training data $\{\{(\mathbf{\Bar{x}}_j,\ \mathbf{w}_j)\}_{j=1}^m\}^N$
\State \# Training stage
\For{$i=1$ to $r$}
    \For{$j=1$ to $m$}
        \State $v_j^{(i)} \gets \operatorname{MLE}\bigl(\{\mathbf{w}_{ij}^{(1)},\  \mathbf{w}_{ij}^{(2)},\ \cdots,\ \mathbf{w}_{ij}^{(N)}\}\bigr)$ \Comment{$\mathbf{w}_{ij}$ is the $i$th dimension of the $j$th noise}
    \EndFor
    \State Get a set $\{(\mathbf{\Bar{x}}_1^{(i)}, v_1^{(i)}),(\mathbf{\Bar{x}}_2^{(i)}, v_2^{(i)}),\cdots, (\mathbf{\Bar{x}}_m^{(i)},v_m^{(i)})\}$
    \State Train a GP to predict variance $\hat{v}_i$ for $i$th dimension
\EndFor
\State \# Prediction stage
\For{$i=1$ to $r$} \\
    \hspace{4mm} $\hat{v}_i \gets GP_i(\mathbf{x}^{(i)})$ 
\EndFor
\State Set $\mathbf{W(x)} = \mathrm{diag}([\hat{v}_1,\ \hat{v}_2,\ \dots,\ \hat{v}_r])$
\State \Return $q(\mathbf{x})\sim \mathcal{N}(\mathbf{0, W(x)})$
% \State Put together the variances of all dimensions into a diagonal covariance matrix, given a new state $\mathbf{x}$, predict $q(\mathbf{x})$, $q(\mathbf{x})\gets GP_q(\mathbf{x})$
\end{algorithmic}
\end{algorithm}

\subsubsection{Estimating KL Divergence Bound} \label{nonstad}
If the true distribution of the noise is non-stationary and varies with the system's state, we need to estimate the joint reference distribution for each horizon, resulting in a varying KL divergence bound $d$ for different horizons. The goal is to find a global maximum $d$ that can bound the ambiguity sets for all horizons. To achieve this, we first combine $n+1$ noise samples from the training data as a joint vector $\mathbf{w}_{(1:n+1)}$, where $n$ is the horizon steps. Subsequently we obtain the joint reference distribution $q_{(1:n+1)}$ using MLE. Next, we draw $M$ samples from this joint reference distribution, each sample being a joint vector of dimension $c\times (n+1)$, where $c$ is the dimension of one noise sample. Then we employ kNN estimation with $N$ true samples of joint noise vectors to get a KL divergence bound for one entire horizon. We then recede the horizon, and the joint noise vector becomes $\mathbf{w}_{(2:n+2)}$, allowing us to obtain another $d$. We repeat this process for each horizon until we have $m-n$ KL divergence bounds, where $m$ is the number of noise samples in the observed data. Finally, we take the maximum of these bounds to obtain the estimated global maximum bound of the KL divergence, as outlined in Algorithm \ref{algo2}.

\begin{algorithm}
\caption{Estimation of global maximum KL divergence bound}\label{algo2}
\begin{algorithmic}[1]
\Require Training data $\{\{(\mathbf{\Bar{x}}_j,\ \mathbf{w}_j)\}_{j=1}^m\}^N$, receding horizon steps $n$, number of samples $M$ drawn from joint reference distribution 
\State Set $d_{max}=0$
\For{$j=1$ to $m-n$}
    \State Combine $n$ noise samples as a joint vector: $\mathbf{w}_{(j:j+n)}=[\mathbf{w}_j,\ \dots,\ \mathbf{w}_{j+n}]^T$
    \State Estimate joint reference distribution: $q_{(j:j+n)}=\operatorname{MLE}(\mathbf{w}_{(j:j+n)})$
    \State Draw $M$ samples of joint vector: $\{\mathbf{\Tilde{w}}_{(j:j+n)}\}^M\sim q_{(j:j+n)}$
    \State Obtain kNN estimation of bound with N true joint noise vectors: $d_j=\operatorname{kNN}\bigl(\{\mathbf{w}_{(j:j+n)}\}^N,\ \{\mathbf{\Tilde{w}}_{(j:j+n)}\}^M\bigr)$
    \If{$d_{j}>d_{max}$}
        \State $d_{max}=d_j$
    \EndIf
\EndFor
\State \Return $d_{max}$
\end{algorithmic}
\end{algorithm}

\section{Experiments}
To validate \ours, we consider a navigation problem with a standard car-like robot \cite{de2005feedback}, which is a widely used model due to its simplicity and effectiveness. The state of the robot is represented by a vector $\mathbf{x}=[x,\ y,\ \theta,\ v]$, where $x$ and $y$ are the coordinates of the robot, $\theta$ is the yaw angle which represents the angle between the orientation of the car and the $x$-axis, and $v$ is the velocity. The control inputs for the robot are the acceleration $a$ and steering angle $\delta$. Our goal is to design a control policy that brings the robot to the origin as close as possible while accounting for the unknown noise.
% \begin{equation}
% \begin{split}
%     \Dot{x} &= vcos(\theta) \\
%     \Dot{y} &= vsin(\theta) \\
%     \Dot{\theta} &= \frac{vtan(\delta)}{L} \\
%     \Dot{v} &= a
% \end{split}
% \end{equation}
The length of the car-like robot is $0.3\ m$. We discretize the system with $dt=0.1\ s$ and use horizon step $n=10$. 

We begin by generating training data through uniform discretization of the state space that the robot can traverse. This process results in a set of states $\{\mathbf{\Bar{x}}_1,\ \mathbf{\Bar{x}}_2,\  \cdots,\ \mathbf{\Bar{x}}_m\}$. For each state $\mathbf{\Bar{x}}_j$, we apply a known control signal, which allows us to obtain the true next state. By subtracting the noiseless next state generated by the dynamical model, we obtain the corresponding true state-dependent noise $\mathbf{w}_j$. We repeat this process $N$ times, resulting in the training data $\{{(\mathbf{\Bar{x}}_j,\  \mathbf{w}j)}_{j=1}^m\}^N$ with $m=1000$ and $N=1000$.

\subsection{True Noise Distribution Construction}
Suppose the true distribution $p$ is a mixture of Gaussians with $h$ finite components: 
\begin{equation}
\begin{split}
    p(\mathbf{x}) &= \sum_i^h \pi_i \mathcal{N}_i(\mathbf{0},\mathbf{W}(\mathbf{x}))\ , \\
    \sum_i^h \pi_i &= 1\ ,
\end{split}
\end{equation}
where each component is a multivariate normal distribution with zero mean and covariance matrix dependent on state $\mathbf{x}$: 
\begin{equation}
    \mathbf{W}_i(x) = diag([{}_{i}\sigma^2_x,\ {}_{i}\sigma^2_y,\ {}_{i}\sigma^2_\theta,\ {}_{i}\sigma^2_v)], \hspace{2mm} i = 1,\ 2,\ \cdots,\ h.
\end{equation}

In the experiments, we consider three different cases of the true noise distribution $p(\mathbf{x})$ as Gaussian mixtures: (a) two-component, (b) three-component, and (c) four-component. The formulas of these true distributions are given in Appendix \footnote{\url{https://github.com/ruiiu/DROC_Variance_Formula}}.

\begin{figure*}[ht]
     \centering
     \begin{subfigure}[b]{0.32\textwidth}
         \centering
         \includegraphics[width=\textwidth]{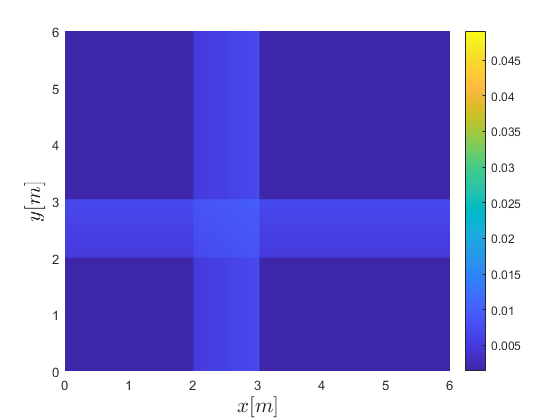}
         \caption{The first component}
         \label{fig:var1}
     \end{subfigure}
     \hfill
     \begin{subfigure}[b]{0.32\textwidth}
         \centering
         \includegraphics[width=\textwidth]{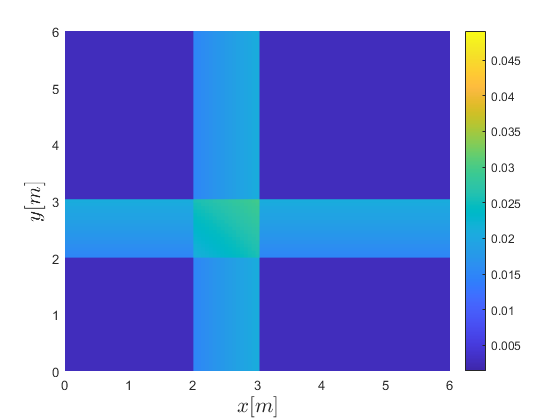}
         \caption{The second component}
         \label{fig:var2}
     \end{subfigure}
     \hfill
     \begin{subfigure}[b]{0.32\textwidth}
         \centering
         \includegraphics[width=\textwidth]{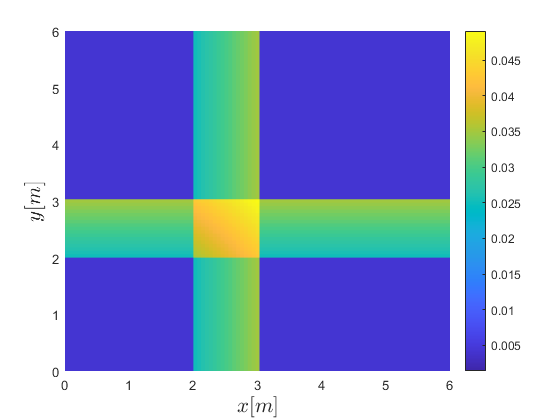}
         \caption{The third component}
         \label{fig:var3}
     \end{subfigure}
        \caption{The heatmap of variances of each component of $p^{(b)}$ for the $x$ and $y$ coordinates.}
        \label{fig:var}
\end{figure*}

Taking the two-component Gaussian mixtures $p^{(b)}$ for illustration, the heatmap of variances of each component of $p^{(b)}$ for the $x$ and $y$ coordinates are shown in Fig. \ref{fig:var}. As we can see, the variances in $x, \ y$ coordinates are larger in the region $2.0\leq x \leq 3.0 \ m$ and $2.0 \leq y \leq 3.0 \ m$. The variances do not change in $\theta$ and $v$ coordinates in the experiments. However, we note that the proposed approach is capable of handling cases where the variances may change in all dimensions of the states.

\subsection{Reference Distribution Estimation}
\begin{figure}[hbt!]
\centering
\begin{subfigure}{.493\linewidth}
  \includegraphics[width=\linewidth]{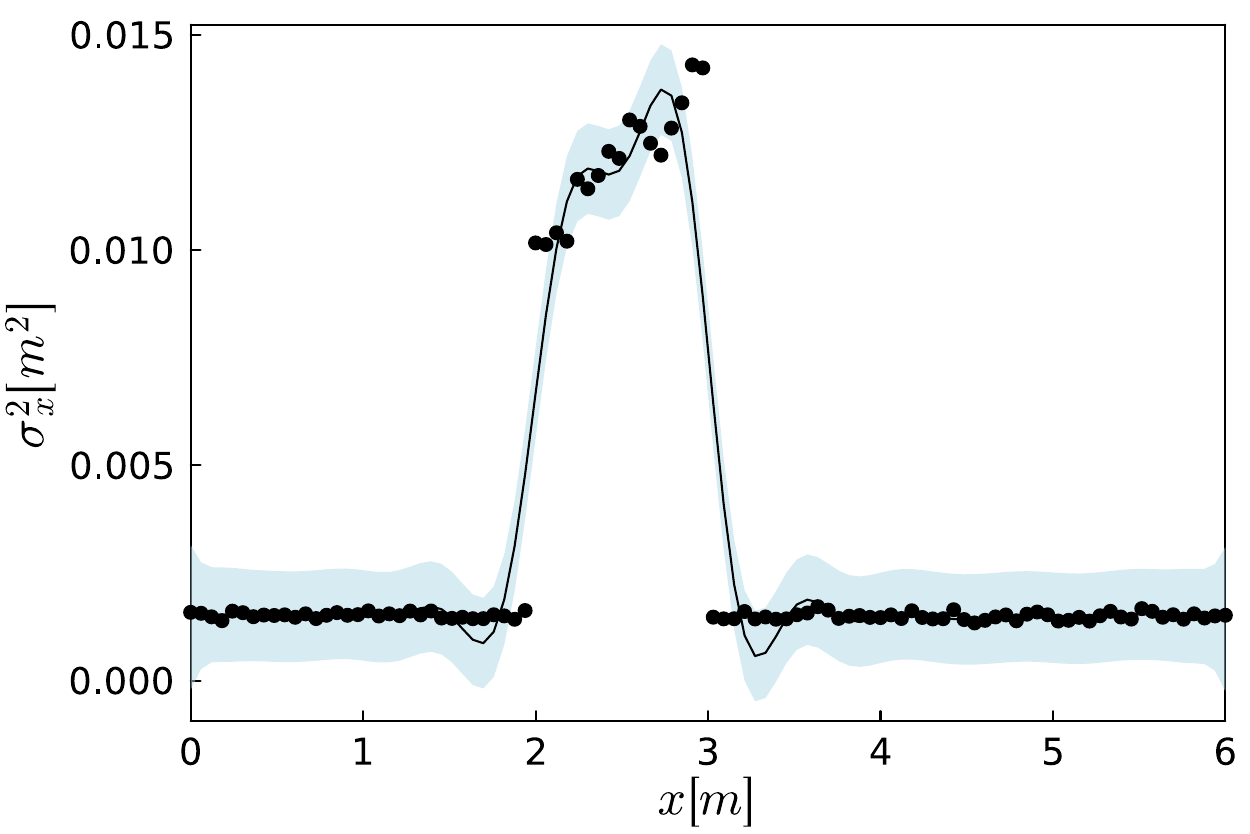}
  \caption{GP prediction for the variance of the $x$ coordinate}
  \label{varx}
\end{subfigure}
\hfill 
\begin{subfigure}{.493\linewidth}
  \includegraphics[width=\linewidth]{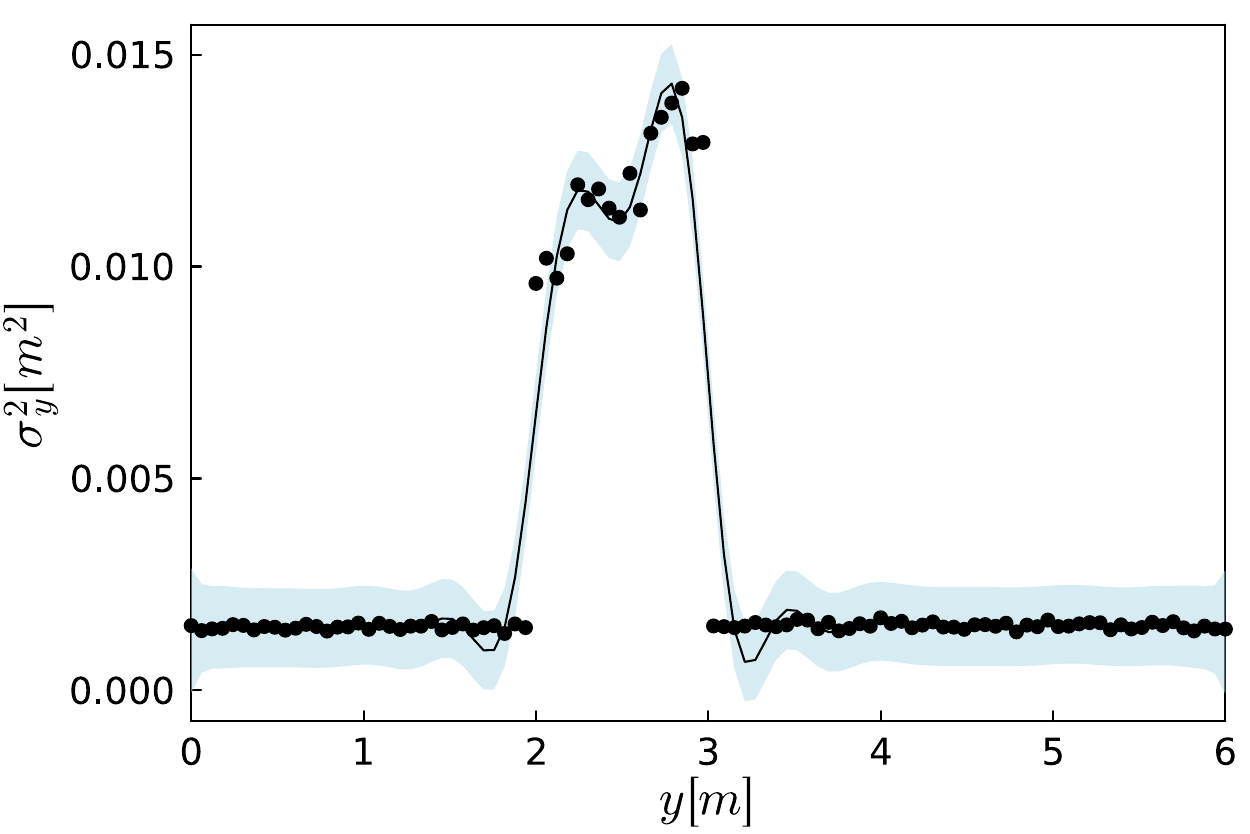}
  \caption{GP prediction for the variance of the $y$ coordinate}
  \label{vary}
\end{subfigure}

\medskip % create some *vertical* separation between the graphs
\begin{subfigure}{.493\linewidth}
  \includegraphics[width=\linewidth]{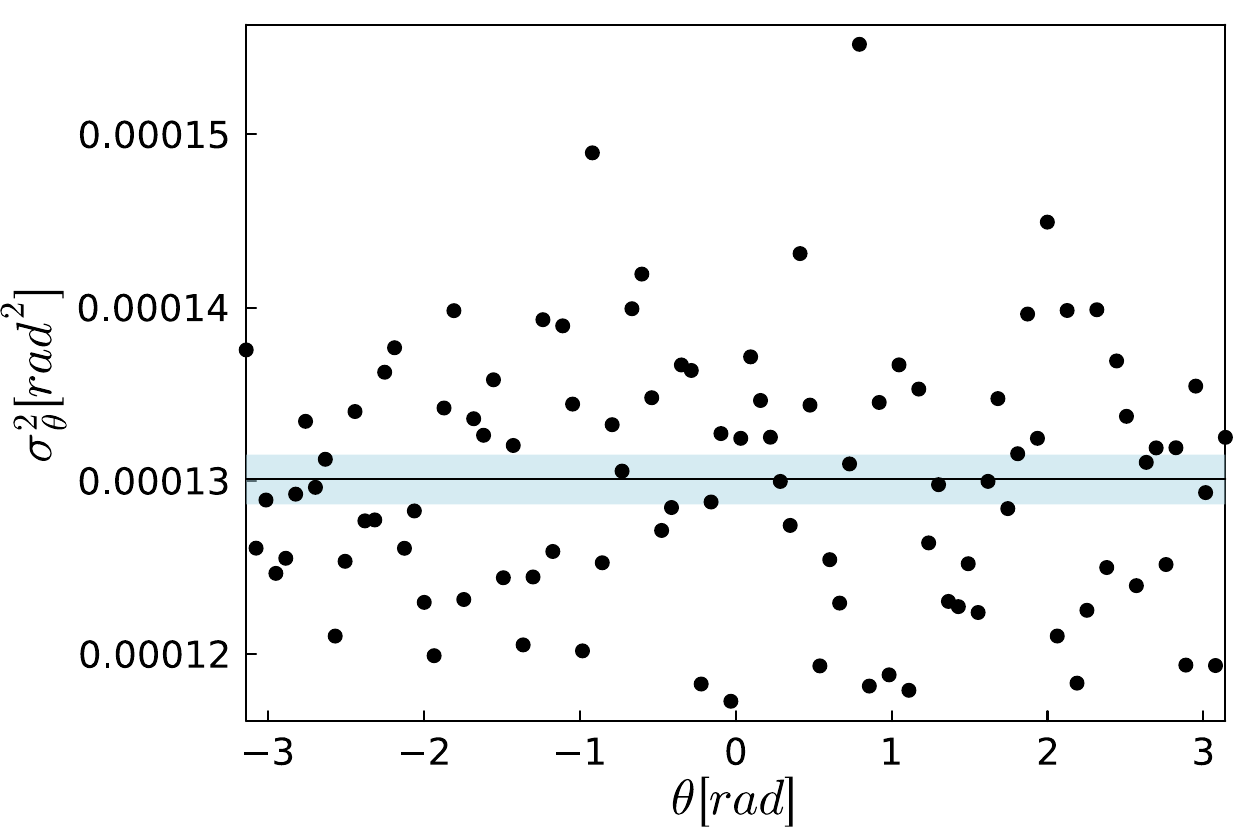}
  \caption{GP prediction for the variance of the $\theta$ coordinate}
  \label{vartheta}
\end{subfigure}
\hfill 
\begin{subfigure}{.493\linewidth}
  \includegraphics[width=\linewidth]{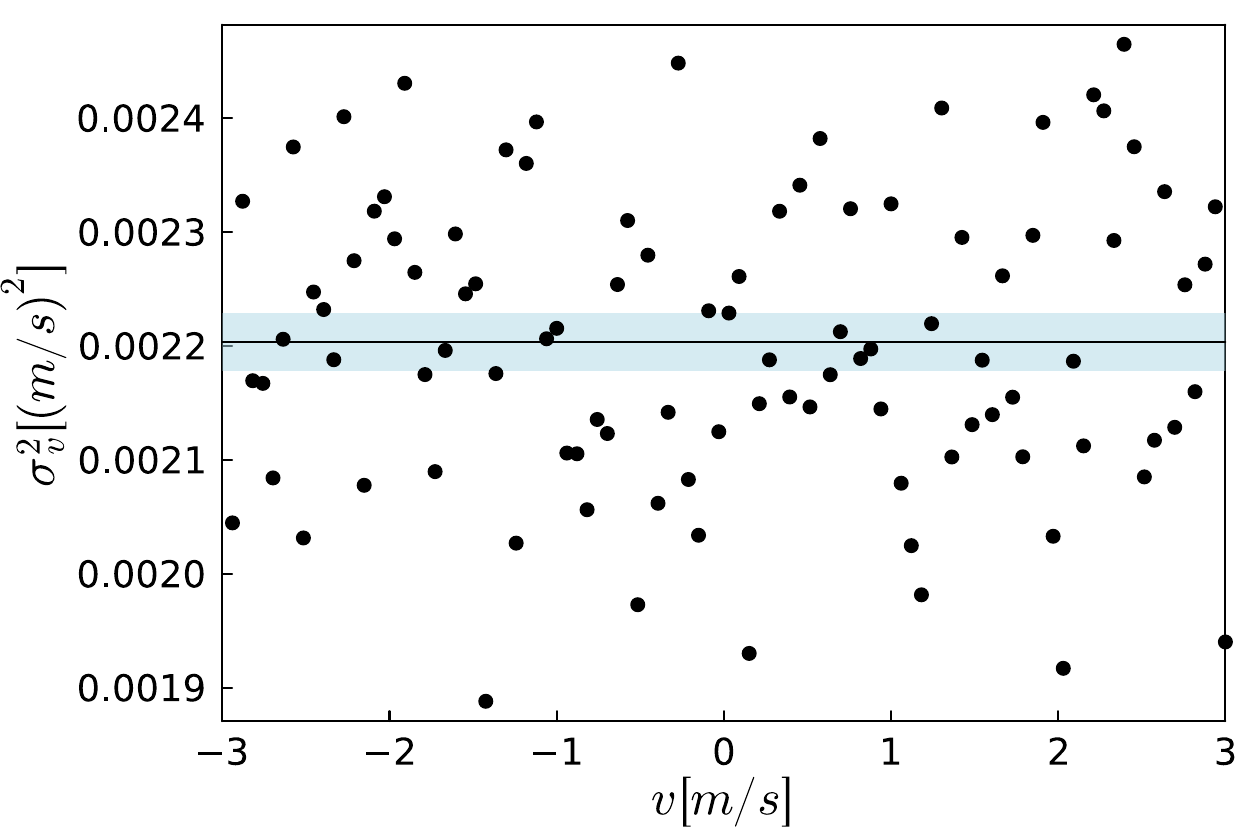}
  \caption{GP prediction for the variance of the $v$ coordinate}
  \label{varv}
\end{subfigure}
\caption{Visualization of the GP used to predict variances for $p^{(b)}$, based on $100$ data points, with black dots representing the variances calculated using MLE from observed data, the line depicting the fitted mean, and the shaded area indicating the $95\%$ confidence interval.}
\label{fig:vargp}
\end{figure}

Considering the three different cases of state-dependent noise described above, we trained a Gaussian process (GP) for each dimension of the state to model the reference distribution $q$, as described in Section \ref{nonstaq}. The GP training results of predicting the variances of $p^{(b)}$ are shown in Fig. \ref{fig:vargp} for instance. The black dots represent the variances estimated using MLE for the $x$ and $y$ coordinates. The line represents the mean, and the shaded area represents the $95\%$ confidence interval. As shown in Fig. \ref{fig:vargp}, the estimated variances in the $x$ and $y$ coordinates are larger in the region $2.0\leq x \leq 3.0 \ m$ and $2.0 \leq y \leq 3.0 \ m$, which captures the distribution of the true noise $p^{(b)}$. In this specific example, the variances do not change in the $\theta$ and $v$ coordinates. The GP-estimated variances in the $\theta$ and $v$ coordinates are ${}^{(b)}\sigma^2_\theta = 1.3e^{-4}$ and ${}^{(b)}\sigma^2_v = 2.2e^{-3}$, respectively. Combining the variances of each dimension into a diagonal covariance matrix allows us to obtain the state-dependent reference distribution $q$, and the results verify our proposed approach.

% \begin{figure}[ht]
%     \centering
%     \subfloat[GP prediction of variance of $p^{(b)}$ for the $x$ coordinate]{
%     {
%     \includegraphics[width=\linewidth]{gp1.pdf}
%     } \label{gp1}}
%     \\
%     \subfloat[GP prediction of variance of $p^{(b)}$ for the $y$ coordinate]{
%     {
%     \includegraphics[width=\linewidth]{gp2.pdf}
%     } \label{gp2}}
%     \caption{Visualization of the Gaussian process used to predict variances for the $x$ and $y$ coordinates of $p^{(b)}$, based on $100$ data points. (a) The GP prediction for the variance of the $x$ coordinate is shown, with black dots representing the variances calculated using maximum likelihood estimation (MLE) from observed data, the line depicting the fitted mean, and the shaded area indicating the $95\%$ confidence interval. (b) The GP prediction for the variance of the $y$ coordinate is shown, with black dots representing the variances calculated using MLE from observed data, the line depicting the fitted mean, and the shaded area indicating the $95\%$ confidence interval.}
%     \label{fig:gp}
% \end{figure}

\subsection{KL Divergence Bound Estimation}
In the example of state-dependent noise, we estimated the KL divergence bound $d$ using the kNN method. For each horizon, we drew $M=100$ samples from the estimated joint reference distribution and used $k=10$ for the kNN estimation. As a result, we obtained KL divergence bound estimates for three different true noise distributions: $d^{(a)}=5.65$ for $p^{(a)}$, $d^{(b)}= 7.28$ for $p^{(b)}$, and $d^{(c)}= 6.22$ for $p^{(c)}$.

\subsection{Comparison with iLQG}
% Once the reference distribution $q$ and KL divergence bound $d$ have been estimated using our proposed data-driven approaches, we can implement the \ours. We use the implementation of DROC based on \cite{nishimura2021rat} with Julia. 

% where the outer loop minimization over the risk-sensitive parameter $\theta$ is carried out using the cross-entropy method. Further details about the implementation can be found in \cite{nishimura2021rat}.

\begin{figure}[ht]
    \centering
    \includegraphics[width=\linewidth]{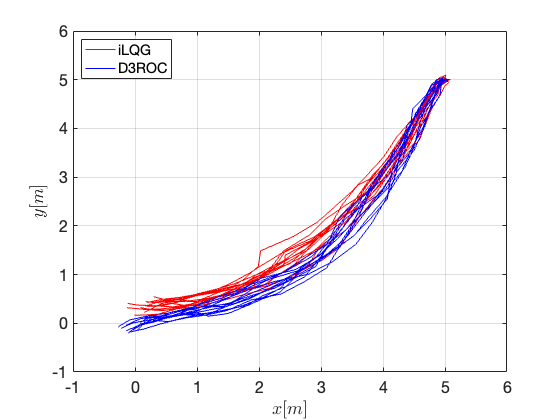}
    \caption{Comparison of a car-like robot navigating to the origin under unknown true noise $p^{(b)}$ with \ours and iLQG. The starting position is $(5.0, 5.0)$ and the goal position is $(0.0,0.0)$. The results of multiple runs are shown in the figure, where the red lines and blue lines represent different paths of the robot under iLQG and \ours, respectively.}
    \label{fig:xycompare}
\end{figure}

We compared the performance of \ours~with the risk-neutral control approach iLQG \cite{athans1971role}. The robot's initial state was $x=5.0\ m$, $y=5.0\ m$ with a yaw angle of $-0.75\pi$ and velocity of zero. The objective was to navigate to the origin under unknown noise. Both control approaches were executed as a MPC for 22 iterations. To ensure accuracy, we performed 15 runs for each true noise distribution, and the resulting paths of the robot under noise $p^{(b)}$ are plotted in Fig. \ref{fig:xycompare}. Both control approaches enable the robot to approach the origin. However, \ours~outperforms iLQG with smaller distances to the origin and more compact paths as the robot gets closer. The plot also reveals \ours's risk-averse behavior, leading the robot to navigate around the region $2.0 \leq x \leq 3.0\ m $ and $2.0 \leq y \leq 3.0\ m $ to avoid areas with higher noise variance. In contrast, iLQG results in more divergent paths, passing through the region $2.0 \leq x \leq 3.0\ m $ and $2.0 \leq y \leq 3.0\ m $.

Additionally, we present the mean distance between the final position of the robot and the origin for multiple runs under different true noise distributions $p$ in Table \ref{comp-table}. It is evident that the mean final distances and their corresponding standard deviations are smaller under \ours~compared to iLQG. This finding further supports the effectiveness of our proposed data-driven approach, \ours, and demonstrates its ability to successfully handle various true noise distributions.

% \begin{table}[ht]
%     \centering
%     \begin{tabular}{|c|c|c|} 
%      \hline
%      Runs & iLQG [m] & DROC [m] \\ 
%      \hline\hline
%      1 & 1.19 & 0.54 \\ 

%      2 & 1.06 & 0.56 \\

%      3 & 1.22 & 0.42 \\

%      4 & 1.20 & 0.39 \\

%      5 & 1.17 & 0.40 \\

%      6 & 1.17 & 0.49 \\

%      7 & 1.15 & 0.47 \\

%      8 & 1.14 & 0.44 \\

%      9 & 1.06 & 0.49 \\

%      10 & 1.19 & 0.37 \\
%     \hline 
%      Mean & \textbf{1.155} & \textbf{0.457} \\
%      Std & 0.055 & 0.064 \\
%      \hline
%     \end{tabular}
%     \caption{The distance between the final position of the robot and the origin of multiple runs under iLQG and DROC}
%     \label{comp-table}
% \end{table}

\begin{table}[ht]
\centering
\resizebox{\linewidth}{!}{%
\begin{tabular}{|l|ll|ll|ll|}
\hline
& \multicolumn{2}{c|}{$p^{(a)}$} & \multicolumn{2}{c|}{$p^{(b)}$} & \multicolumn{2}{c|}{$p^{(c)}$} \\ \hline
Distance & \multicolumn{1}{c|}{iLQG{[m]}} & \ours{[m]} & \multicolumn{1}{c|}{iLQG{[m]}} & \ours{[m]} & \multicolumn{1}{c|}{iLQG{[m]}} & \ours{[m]} \\ \hline\hline
Mean & \multicolumn{1}{c|}{\textbf{0.37}} & \multicolumn{1}{c|}{\textbf{0.25}} & \multicolumn{1}{c|}{\textbf{0.36}} & \multicolumn{1}{c|}{\textbf{0.17}} & \multicolumn{1}{c|}{\textbf{0.33}} & \multicolumn{1}{c|}{\textbf{0.28}} \\
Std & \multicolumn{1}{c|}{0.14} & \multicolumn{1}{c|}{0.08} & \multicolumn{1}{c|}{0.11} & \multicolumn{1}{c|}{0.09} & \multicolumn{1}{c|}{0.15} & \multicolumn{1}{c|}{0.10} \\ \hline
\end{tabular}
}
\caption{The mean distance between the final position of the robot and the origin, along with its standard deviation, for multiple runs under iLQG and \ours. The experiments were conducted using three different true noise distributions.}
\label{comp-table}
\end{table}

\section{Conclusions}
In conclusion, this paper presents \ours, a data-driven approach that overcomes the limitation of traditional DROC methods requiring known ambiguity sets for noise distribution. We evaluate our approach through a navigation problem for a car-like robot with unknown noise distributions. The numerical results demonstrate that \ours~achieves smaller mean final distances between the robot and the origin, along with lower corresponding standard deviations, compared to iLQG. Additionally, the risk-averse behavior of \ours~enables the robot to tend to avoid regions with higher noise variance, whereas iLQG leads to paths passing through such regions. Furthermore, our approach proves effective in handling various noise distributions. Overall, \ours~offers a promising solution to real-world DROC problems where noise distribution and KL divergence bound are unknown, making the DROC framework more practical and applicable.

\bibliographystyle{ieeetr}
\bibliography{reference}

\end{document}